\documentclass[letterpaper, 10 pt, conference]{ieeeconf}  

\IEEEoverridecommandlockouts                              

\usepackage{cite}
\usepackage{amsmath,amssymb,amsfonts}
\usepackage{graphicx}
\usepackage{textcomp}
\usepackage{xcolor}

\usepackage{algorithm}
\usepackage{algorithmicx}  
\usepackage{amsfonts}
\usepackage{amsmath}
\usepackage{amsthm}
\usepackage{adjustbox}
\usepackage{bm}
\usepackage{booktabs}
\usepackage{diagbox}
\usepackage{dsfont} 
\let\labelindent\relax  
\usepackage{enumitem}
\usepackage{etoolbox}
\usepackage{hyperref}  
\usepackage{lipsum}  
\usepackage{mathtools}
\usepackage{multirow}
\usepackage{nicefrac}
\usepackage[final]{pdfpages}
\usepackage{siunitx}
\usepackage{stfloats}  
\usepackage[caption=false]{subfig}
\usepackage{tabu}
\usepackage{threeparttable}  
\usepackage{tikz}
\usepackage{tikzscale}
\usepackage{xcolor}
\usepackage{xfrac}

\usepackage{pgfplots}   
\pgfplotsset{compat=newest}
\usepgfplotslibrary{groupplots}
\usetikzlibrary{backgrounds}
\usetikzlibrary{arrows.meta}
\usetikzlibrary{calc,positioning}
\usetikzlibrary{3d}

\graphicspath{{img/}}
\hypersetup{colorlinks=false}

\definecolor{githubColor}{HTML}{2EA44F}
\newcommand{\gitref}[2]{\href{#1}{\color{githubColor}{#2}}}%

\definecolor{newGray}{HTML}{808080}

\definecolor{matlabYellow}{rgb}{0.9290, 0.6940, 0.1250}%
\definecolor{matlabPurple}{rgb}{0.9290, 0.6940, 0.1250}%
\definecolor{matlabLBlue}{rgb}{0.3010, 0.7450, 0.9330}%
\definecolor{matlabGreen}{rgb}{0.4660, 0.6740, 0.1880}%
\definecolor{matlabRed}{rgb}{0.8500, 0.3250, 0.0980}%
\definecolor{matlabBlue}{rgb}{0, 0.4470, 0.7410}%

\definecolor{colorCircle}{HTML}{0072BD}

\definecolor{colorRect}{HTML}{D95319}

\newcolumntype{O}[1]{S[detect-weight, mode=text, table-format=#1]}

\renewcommand{\bfseries}{\fontseries{b}\selectfont} 
\robustify\bfseries             
\newrobustcmd{\B}{\bfseries}

\newcommand\copyrighttext{\footnotesize \textcopyright~2023 IEEE. Personal use of this material is permitted. Permission from IEEE must be obtained for all other uses, in any current or future media, including reprinting/republishing this material for advertising or promotional purposes, creating new collective works, for resale or redistribution to servers or lists, or reuse of any copyrighted component of this work in other works.%
}

\newcommand\copyrightnotice{%
    \begin{tikzpicture}[remember picture,overlay]%
 	\node[%
        anchor=south, %
        yshift=10pt%
    ] at (current page.south)%
 	{\fbox{\parbox{\dimexpr\textwidth-\fboxsep-\fboxrule\relax}{\copyrighttext}}};%
 	\end{tikzpicture}%
}

\newtheoremstyle{tstyle}
  {}
  {}
  {\itshape}
  {}
  {\bfseries}
  {.}
  { }
  {\thmname{#1}\thmnumber{ #2}\thmnote{ (#3)}}%
\theoremstyle{tstyle}

\newcommand{\comp}[1]{{#1}^{\mathsf{c}}}

\newcommand{\trans}{^\text{\rmfamily \textup{T}}}

\newcommand{\pspace}{\,}  

\title{
Adaptive Patched Grid Mapping
}

\author{Thomas Wodtko, Thomas Griebel, and Michael Buchholz%
\thanks{Parts of this work were supported by the State Ministry of Economic Affairs Baden-Württemberg (project U-Shift\,II, AZ\,3-433.62-DLR/60). Parts of this research have been conducted as part of the EVENTS project, which is funded by the European Union, under grant agreement No 101069614. Views and opinions expressed are however those of the author(s) only and do not necessarily reflect those of the European Union or European Commission. Neither the European Union nor the granting authority can be held responsible for them.}%
\thanks{All authors are with the Institute of Measurement, Control and Microtechnology, Ulm University, Albert-Einstein-Allee 41, 89081 Ulm, Germany {\tt\footnotesize \{firstname\}.\{lastname\}@uni-ulm.de}}%
}

\begin{document}

\maketitle

\begin{abstract}
In this work, we propose a novel adaptive grid mapping approach, the Adaptive Patched Grid Map, which enables a situational aware grid based perception for autonomous vehicles.
Its structure allows a flexible representation of the surrounding unstructured environment.
By splitting types of information into separate layers less memory is allocated when data is unevenly or sporadically available.
However, layers must be resampled during the fusion process to cope with dynamically changing cell sizes.
Therefore, we propose a novel spatial cell fusion approach.
Together with the proposed fusion framework, dynamically changing external requirements, such as cell resolution specifications and horizon targets, are considered.
For our evaluation, real-world data were recorded from an autonomous vehicle driving through various traffic situations.
Based on this, the memory efficiency is compared to other approaches, and fusion execution times are determined.
The results confirm the adaptation to requirement changes and a significant memory usage reduction.
\end{abstract}

\section{INTRODUCTION}
\label{sec:intro}

\copyrightnotice%
Deploying robots with an increasing level of autonomy to cope with a broader range of applications requires an abstract representation of unstructured environments to operate within those~\cite{elfes1989occupancy}.
Today, not only robots in small working environments but also autonomous vehicles rely on grid-based environment representations~\cite{thrun2005probalilistic}.
The main goal of available approaches~\cite{nuss2016random, schreiber2022multitask, richter2022mapping} is to classify a cell's state over time robustly.
For moving objects, velocities are estimated using either classical approaches~\cite{nuss2016random} or deep learning methods~\cite{schreiber2022multitask}.

In general, grid map-based sensor fusion allows for using several sensor types.
Especially, lidar, camera, and radar sensors, most commonly used for autonomous driving, can be integrated.
The Dempster-Shafer theory of evidence (DST)~\cite{pearl1988probabilistic,smets1990combination} is often used to combine different types of information from these sensors.
It enables the combination of binomial information, e.g., if a cell is occupied or free, with multinomial semantic class labels.

However, with a growing operational design domain (ODD) and, thus, an increasing area of interest, grid map-based sensor fusion must meet more challenging demands, such as capturing and fusing data of large areas, while staying computationally efficient.
Therefore, different strategies for managing cells in a grid map format have emerged~\cite{buerkle2020efficient, wellhausen2021efficient}.
By altering the number of cells allocated at a time, these strategies mainly try to reduce computational expenses.
Alternatively, some autonomous vehicles decentralize the fusion process~\cite{ushift2022}.
For this, specific sensor modules acquire and preprocess sensor data before the information is gathered and fused in a central unit.
While the workload is distributed and parallel processing is possible, these architectures require transmitting intermediate results over a link with a limited data rate.
Hence, such a distribution can reduce processing time for the cost of link latency.
\begin{figure}[t]
    \centering
	\includegraphics[width=0.9\columnwidth,]{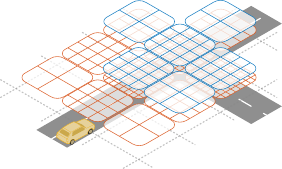}%
    \caption{%
        The APGM structure is illustrated for a vehicle (yellow) driving on the road approaching an intersection.
        The environment is divided into patches (dashed gray), and each patch contains a variable number of layers (colored rounded rectangles).
        Each layer has a variable cell resolution (rasterization). 
        For illustration reasons, only the area in front of the vehicle shows layers.
        The cell resolution is set according to the current scenario. 
        Here, layers with a road have higher cell resolutions than other layers at the same distance. 
    }
    \label{fig:apgm}
\end{figure}
Additionally, recent publications discuss the issue that autonomous vehicles tend to run the entire software stack continuously~\cite{henning2022situation, henning2022situation2}.
For that, software components must adaptively realize situational requirements to become more efficient.

To the best of our knowledge, we propose the first adaptive grid map structure allowing for a dynamic cell resolution and a varying amount of layers, illustrated in Fig.~\ref{fig:apgm}.
With this work, we propose the following:
\begin{itemize}
    \item The Adaptive Patched Grid Map (APGM) in Section~\ref{sec:apgm};
    \item a mathematical derivation of the generic APGM fusion framework~\ref{sec:fusionFrame};
    \item fusion operators for adaptively changing cell resolutions in Section~\ref{sec:adapRes}; and
    \item an open-source C++ library\footnote{\gitref{https://github.com/wodtko/adaptive_patched_gridmap}{https://github.com/wodtko/adaptive\_patched\_gridmap}} of the adaptive grid map and its fusion framework.
\end{itemize}

\section{RELATED WORK}
\label{sec:relWork}

The use of grid maps for perception and navigation tasks originates from robotics~\cite{elfes1989occupancy,thrun2005probalilistic}.
Splitting the robot's environment into cells is an abstract representation that allows for using sensor measurements of an unstructured environment without the need for object detectors.
A classic grid map combines all probabilities for a cell to be occupied over time.
For this, it is assumed that the measurements of different cells are independent.
The grid map size is chosen to cover the whole working environment; thus, the memory allocation stays constant.

However, with a larger working environment, the execution times of grid maps increases.
Especially for autonomous vehicles, it is infeasible to allocate a single grid map covering multiple kilometers with cells on a centimeter scale.
Hence, the first approaches for autonomous vehicles usually create a grid map of only the near field environment to stay computationally efficient~\cite{danescu2011modeling, nuss2013decision}.
Even recent approaches with increased capabilities mostly use this strategy for computational reasons~\cite{nuss2016random, richter2022mapping}.
For the fusion of multiple sensor modalities and to allow classical approaches for velocity, a cell is usually represented by a basic belief assignment of the Dempster-Shafer theory.
Further, in addition to the cell's occupancy state, velocities are estimated in a dynamic occupancy grid map.
For this, either classical~\cite{richter2022mapping} or deep learning~\cite{schreiber2022multitask} approaches are available.

In this work, however, the focus is on improving the underlying grid map structure rather than the dynamic occupancy estimation.
Similarly, recent approaches proposed new grid map structures and analyzed the implementation impact on efficiency~\cite{buerkle2020efficient, wellhausen2021efficient}.
Retaining a moving grid, the authors of~\cite{buerkle2020efficient} proposed a method to reduce the cell resolution for the far field of the environment.
While they can effectively lower the number of cells required, the grid map layout must be set once before execution.
Thus, all possible requirements must be met with a single layout, and requirement changes can not lead to further improvements.
In contrast, the authors of~\cite{wellhausen2021efficient} proposed different ways of dividing the environment into sub-maps first.
In principle, this would allow reacting dynamically on external requirements.
However, the focus of~\cite{wellhausen2021efficient} is to analyze different container structures and implementations concerning access and processing times.
The environment is also divided into sub-maps by~\cite{kichun2018cloud, chansoo2021geodetic}.
While~\cite{chansoo2021geodetic} focuses on the localization within a large-scale environment rather than representing the immediate surroundings,
the authors of~\cite{kichun2018cloud} propose the use of sub-maps in combination with infrastructure sensors.
Efficiently considering infrastructure information is the main benefit of using sub-map grid maps.
Hence, the combination of the adaptive grid map of this work with infrastructure information is part of future work.

Even though some approaches improve efficiency by neglecting some areas~\cite{wellhausen2021efficient} or by choosing non-uniform cell resolutions~\cite{buerkle2020efficient}, all approaches only consider static requirements.
Static, in this case, implies that all requirements are predefined and any requirement change during run time cannot be used for further efficiency improvements.
Recent publications, however, provide situation-aware requirements for modules within autonomous vehicles~\cite{henning2022situation}.
Those requirements comprise resolution demands, areas of interest, and horizon targets.
Since most algorithms in autonomous vehicles do not consider such requirements, the authors propose to deactivate not required modules or to crop their input data to save energy.
This shows the need for approaches taking those requirements into account, allowing further improvements of methods as proposed by~\cite{henning2022situation}. 
In the case of modular, decentralized sensor setups~\cite{ushift2022}, the potential for savings is even higher since certain areas are always unconsidered~\cite{henning2022situation2}. 
To the best of our knowledge, no grid mapping algorithm exists that adaptively realizes dynamic requirements, as provided by~\cite{henning2022situation, henning2022situation2}.

\section{FOUNDATIONS}
\label{sec:foundations}
This section gives a brief overview of the required fundamentals. 
In Section~\ref{sec:funDS} the basics of the DST are outlined and Section~\ref{sec:funGM} briefly compares two fundamental approaches for a grid map based perception.
For more details, additional sources are given in the respective subsections.

\subsection{Dempster-Shafer Theory of Evidence}
\label{sec:funDS}
In contrast to the Bayesian representation of information in grid maps~\cite{elfes1989occupancy}, the DST enables decision-free reasoning by considering the evidence of information.
Therefore, the DST is introduced in this section.
More detailed information can be found in~\cite{pearl1988probabilistic,smets1990combination,sentz2002combination}.

Three elementary functions are of importance in the DST: the basic belief assignment function (BBA) $m$, the belief function $Bel$, and the plausibility function $Pl$.
The universal set which contains all considered elementary states or hypotheses is called the \textit{frame of discernment} $\Omega$.
Later, e.g., for lidar occupancy measurements $\Omega=\{occupied, free\}$.
A BBA defines a mapping of evidence mass to all subsets $A \subset \Omega$.
Formally, a BBA $m : 2^\Omega \rightarrow [0,1]$ is defined by
\begin{subequations}
\label{eq:bbaDefinition}
\begin{align}
\label{eq:bbaDefinitionEmpty}
    m(\emptyset) &= 0 \pspace , \\
    \sum_{A \subseteq 2^\Omega}{m(A)} &= 1 \pspace ,
\end{align}
\end{subequations}
where $2^\Omega$ is the power set of $\Omega$, and $\emptyset$ the empty set.
Since~\eqref{eq:bbaDefinitionEmpty} must hold for all BBAs; it is always implicitly considered in the following to enable better readability.
Therefore, the value of BBAs for the empty set is not explicitly shown.
For $A \subseteq 2^\Omega, |A| \geq 2$, the value of $m(A)$ represents all available evidence mass supporting any but not a specific hypothesis in $A$.
Thus, the uncertainty of a BBA is represented by $m(\Omega)$ as it is the evidence mass that cannot be assigned to any subset.
Similar to~\cite{nuss2013decision}, a BBA is assumed to be a probability function in this work.

Given a BBA, the belief and plausibility of a set $A$ are the upper and lower bound to the interval containing the probability of $A$.
Generally speaking, the belief is the sum of evidence that explicitly supports $A$, while the plausibility is the sum of evidence that $A$ does not entirely contradict.
Therefore, the belief and plausibility can be interpreted as a pessimistic and an optimistic guess of the exact probability of $A$, respectively.
Formally, the belief function $Bel: 2^\Omega \rightarrow [0,1]$ and the plausibility function $Pl: 2^\Omega \rightarrow [0,1]$ are defined by
\begin{subequations}
\label{eq:belPlDefinition}
\begin{align}
    Bel(A) &= \sum_{B \subseteq A}{m(B)} \pspace , \\
    Pl(A) &= \sum_{B \cap A \neq \emptyset}{m(B)} \pspace .
\end{align}
\end{subequations}
With the complement of $A$ denoted by $\comp{A}$, both functions are connected by
\begin{equation}
    Pl(A) = 1 - Bel(\comp A) \pspace .
\end{equation}

In order to incorporate multiple BBAs into the reasoning process, BBAs can be merged using the Dempster-Shafer rule of combination (DST-RC). 
The combination of two BBAs $m_1$ and $m_2$, denoted as $m_{1 \oplus 2}$, is defined by
\begin{subequations}
\begin{gather}
    m_{1 \oplus 2}(A) = \cfrac{ \sum\limits_{B \cap C = A}{m_1(B)m_2(C)} }{ 1 - K} \pspace , \\[2mm]
    \text{with } K = \sum_{B \cap C = \emptyset}{m_1(B)m_2(C)} \pspace .
\end{gather}
\end{subequations}
Here, $K \in (0,1)$ represents the belief mass associated with the conflict between the two BBAs.
The neutral element to the DST-RC is the vacuous BBA with $m(\Omega) = 1$.
As shown in~\cite{smets1990combination,sentz2002combination}, the DST-RC can lead to counter-intuitive results, especially for $|\Omega| > 2$.

Fusing information from multiple cells and, thus, choosing combination operators is of interest in later sections.
In anticipation of later reference, an example of the DST-RC is given for the binomial case.
With a frame of discernment $\Omega = \{A, B\}$ and the two strongly conflicting BBAs $m_1$ and $m_2$ the combined BBA $m_3 = m_{1 \oplus 2}$ is given by
\begin{equation}
\label{eq:dstRuleEx}
\begin{aligned}
    & m_1(A) = 0.9 \pspace , \pspace && m_1(B) = 0   \pspace , \pspace && m_1(\Omega) = 0.1, \\
    & m_2(A) = 0   \pspace , \pspace && m_2(B) = 0.9 \pspace , \pspace && m_2(\Omega) = 0.1, \\
    & m_3(A) \approx 0.47 \pspace , \pspace && m_3(B) \approx 0.47 \pspace , \pspace && m_3(\Omega) \approx 0.05.
\end{aligned}
\end{equation}
Here, the uncertainty decreases by accumulating equally uncertain BBAs using the DST-RC, even in strong conflicts.
This shows that the uncertainty $m_3(\Omega)$ does not represent the consistency of gathered information, but rather the amount of evidence received for a certain state.

Whenever there is the need to make a decision, belief or plausibility can be used; however, the uncertainty information would be ignored.
In contrast, using the pignistic transformation~\cite{smets1990combination}
\begin{equation}
    BetP_m(A) = \sum_{B \subseteq \Omega}{\cfrac{|A \cap B|}{|B|} \, m(B)} \pspace , 
\end{equation}
a BBA $m$ can be transformed into a probabilistic function considering the uncertainty of $m$, which can then be consulted for a decision.

Given a BBA of a source knowing that this source is only reliable to a probability of $\alpha \in [0,1]$, the BBA can be discounted~\cite{huynh2009discounting} to reflect this by
\begin{equation}
    m^{\alpha}(A) = 
    \begin{cases}
    \begin{aligned}
        1 - \alpha \pspace + \pspace & \alpha \cdot m(\Omega) \pspace , & \quad \text{if } A  = \Omega \pspace , \\
                     & \alpha \cdot m(A) \pspace , & \quad \text{else} \pspace . 
    \end{aligned}
    \end{cases}
\end{equation}

\subsection{Grid Mapping}
\label{sec:funGM}
The basic idea of grid maps is to divide an unstructured environment into cells. 
Each cell contains information about the state of the location represented by the respective cell. 
The type of information is task-specific, e.g., a cell can contain simple probabilities for a cell being occupied~\cite{elfes1989occupancy} or multinomial information including semantic labels and occupancy evidences~\cite{richter2022mapping}.
In most state-of-the-art approaches for autonomous vehicles~\cite{nuss2016random, schreiber2022multitask, richter2022mapping}, BBAs are used to encode information, allowing efficient processing.
When using BBAs, the dimension of the \textit{frame of discernment} $\Omega$ defines the amount of bytes required to represent a cell and, thus, the required memory space and link latency when storing or transmitting grid map information respectively.

\begin{figure}
    \centering
    \subfloat[Sub-Map Strategy]{%
        \label{fig:patchedVScenter:patched}%
    	\includegraphics[
    	    width=0.48\columnwidth,
    	]{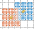}%
    }
    \hfill
    \subfloat[Moving Strategy]{%
        \label{fig:patchedVScenter:center}%
    	\includegraphics[
    	    width=0.48\columnwidth,
    	]{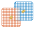}%
    }
    \caption{%
        Two major strategies to store grid map cells for autonomous vehicles are shown. 
        In~\protect\subref{fig:patchedVScenter:patched} the world is divided into sub-maps (dashed gray), which may contain cells (rasterized color), instead in~\protect\subref{fig:patchedVScenter:center} cells are stored in a single rectangular grid moving with the vehicle.
        Red cells illustrate the grid map for the lower left vehicle (yellow) and the blue cells for the upper right vehicle (yellow).
    }
    \label{fig:patchedVScenter}
\end{figure}

For the cell division of the environment, two major dividing strategies require different memory interactions.
Fig.~\ref{fig:patchedVScenter} illustrates both strategies for a moving autonomous vehicle.
The currently most common strategy for autonomous vehicles is to have a single grid map with a constant cell pattern~\cite{schreiber2022multitask, richter2022mapping}.
When the vehicle moves, the grid map is shifted with the vehicle, keeping it in the center of the map. 
By this, the amount and pattern of cells are kept constant, allowing storage of all cells in a fixed layout.

In contrast, the environment can also be divided into sub-maps first.
Each sub-map is referenced to a global datum and is not moved with the vehicle.
The grid cells are then stored within these sub-maps.
Therefore, sub-maps need to be dynamically created or deleted concerning the vehicle position, which increases the complexity of storing cell data.

The main advantage of using a single-moving grid is computational efficiency.
On the other hand, using a sub-map strategy allows for allocating sub-maps only where they are required~\cite{wellhausen2021efficient}.

\section{METHOD}
\label{sec:method}
The APGM is presented in this section. 
First, the general structure is outlined, and a mathematical representation is given.
Afterward, the issue of fusing differently sized cells is explained and a fitting fusion operator is presented.
It shall be mentioned, that the derivation of the fusion operator is focusing on points cloud measurements of, e.g., lidar sensors, using the measurement model of~\cite{richter2022mapping}. 
Although a fusion operator for camera semantic labels can be derived in a similar manner, its definition for an arbitrary amount of  semantic hypotheses is part of future work.

\subsection{Adaptive Patched Grid Map}
\label{sec:apgm}
Here, the APGM is textually described first; then, a precise definition is given.
The APGM is a patched grid map with a dynamic number of layers per patch.
It enables the realization of external requirements, especially resolution demands are considered.

As illustrated in Fig.~\ref{fig:apgm}, the APGM uses the sub-map strategy to divide the environment into sub-maps, called patches in this work.
Similar to~\cite{wellhausen2021efficient}, a grid map has a global reference datum, a patch edge length, and a dynamic set of patches.
Instead of directly dividing each patch into cells, a patch contains a dynamic set of layers.
Each layer then divides the environment into cells.
Therefore, areas in the environment can be covered by multiple cells of different layers.
In contrast to a cell containing all conceivable information~\cite{richter2022mapping}, the information is split into smaller portions using separate layers.
The resolution of a layer can be individually defined, meaning that layers of the same patch can have different resolutions.
Furthermore, two layers of the same type in different patches may also have different resolutions.

Data availability can influence the grid map layout by dividing cell information into different layers.
When, e.g., two sensor types with different fields of view (FOV) are used, each layer must only be available within the FOV of the respective sensor type.
Especially when sensors are temporarily switched off, respective layers do not need to be allocated.
Further, due to the sub-map strategy, specific areas can be omitted entirely whenever external requirements allow.
By this, layers and the corresponding cells can be placed and allocated according to dynamic external requirements without redefining the whole structure.

\begin{figure}[t]
\vspace{0.1cm}
    \centering
    \def\svgwidth{\linewidth}
    \input{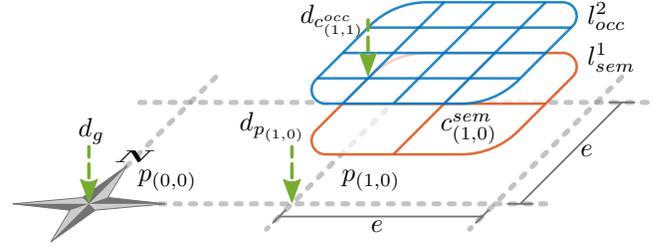}
    \caption{%
        A grid map $g$ with two patches, $p_{(0,0)}$ and $p_{(1,0)}$, the edge length $e$, and the global reference $d_g$ is illustrated.
        The patch $p_{(0,0)}$ is empty and the patch $p_{(1,0)}$ with reference $d_{p_{(1,0)}}$ contains two layers, $l^2_{occ}$ and $l^1_{sem}$, with $T = \{occupancy (occ), semantic (sem)\}$.
        The cell $c^{sem}_{(1,0)}$ is a cell of the semantic layer with index $(1,0)$ and $d_{c^{occ}_{(1,1)}}$ the reference of a cell of the occupancy layer with index $(1,1)$.
    }
    \label{fig:mathIllu}
\end{figure}

In the following, a precise definition of the APGM is given, and an exemplary visual reference to the symbols used is illustrated in Fig.~\ref{fig:mathIllu}.
The structure is defined bottom up for better readability, starting on the cell level first.
The sensor measurement information is split into parts; each part has a specific type $t \in T$, where $T$ is the set containing all information types with the cardinality $|T| > 0$.
For example, $T = \{occupancy (occ), semantic (sem), velocity (vel)\}$.
Further, $\mathcal{C}_t$ is the value space for cells $c^t \in \mathcal{C}_t$ of type $t$.
For type $velocity$ $\mathcal{C}_{vel} = \mathbb{R}^2$.
A layer $l_t^r \in \mathcal{L}$ with resolution step $r \in \mathbb{N}$ containing cells of type $t$ is defined as
\begin{align}
\label{eq:defLayer}
    l_t^r \in \mathcal{C}_t^{\pspace m \times m} \pspace , \pspace m = 2^r
\end{align}
and the corresponding set $\mathcal{L}$ containing all layers is given by
\begin{align}
    \mathcal{L} = \{l_t^r\}_{t \in T, r \in \mathbb{N}} \pspace .
\end{align}
Using powers of two for the dimension $m \in \mathbb{N}$ allows a more effortless fusion and resampling process; generally, an arbitrary resolution could be set.
Next, each patch $p_{(i_x,i_y)} \in \mathcal{P}$ has an index $(i_x,i_y) \in \mathbb{Z} \times \mathbb{Z}$ and is a set of layers, i.e.,
\begin{align}
    p_{(i_x, i_y)} \subset \mathcal{L} \pspace .
\end{align}
At most, one layer may be in a patch for each type $t \in T$, which means
\begin{align}
    \forall l_{t_j}^{r_j}, l_{t_k}^{r_k} \in p_{(i_x,i_y)}: j \neq k \Rightarrow t_j \neq t_k \pspace . 
\end{align}
The set $\mathcal{P}$ containing all patches is defined by
\begin{subequations}
\begin{gather}
    \mathcal{P} = \{p_{(i_x,i_y)}\}_{(i_x,i_y) \in \mathbb{Z} \times \mathbb{Z}} \pspace .
\end{gather}
\end{subequations}
Given a patch $p \in \mathcal{P}$, the index set for this patch $\mathcal{I}_p \subset T$, containing all types for which a layer is available in $p$, is defined by
\begin{align}
    \mathcal{I}_p = \{ t \in T : \pspace \exists l_t^r \in p\} \pspace .
\end{align}
Finally, an APGM $g \in \mathcal{G}$ is a set containing patches.
Each index has at most one patch in a grid map.
The grid map $g$ and the set $\mathcal{G}$ are formally given by
\begin{align}
    g \subset \mathcal{P} \qquad \text{and} \qquad \mathcal{G} = 2^\mathcal{P} \pspace .
\end{align}
Given a grid map $g \in \mathcal{G}$, the index set for this grid map $\mathcal{I}_g \subset \mathbb{Z} \times \mathbb{Z}$, containing all indices for which a patch is available in $g$, is defined by
\begin{align}
    \mathcal{I}_g = \{ (i_x,i_y) \in \mathbb{Z} \times \mathbb{Z} : \pspace \exists p_{(i_x,i_y)} \in g\} \pspace .
\end{align}

To reference a patch or a cell to a specific area in the environment, the edge length of patches and a geodetic reference of the grid map must be specified.
Keeping them constant during execution simplifies the fusion process.
In our case, the grid map is referenced to the UTM~\cite{snyder1987map} origin.
Given the patch index $(i_x,i_y)$, the edge length of patches $e \in \mathbb{R}$ and the grid map $g$ with reference $d_g \in \mathbb{R}^2$, the reference  of a patch $d_{p_{(i_x,i_y)}} \in \mathbb{R}^2$ can be calculated by
\begin{align}
    d_{p_{(i_x,i_y)}} = d_g + \begin{bmatrix}e \cdot i_x & e \cdot i_y\end{bmatrix}\trans \pspace .
\end{align}
Subsequently, the reference $d_{c^t_{(a,b)}} \in \mathbb{R}^2$ of a cell $c^t_{(a,b)} \in l_t^r \in p$ within the patch $p$ is given by
\begin{align}
    d_{c^t_{(a,b)}} = d_p + e / 2^r\begin{bmatrix}a & b\end{bmatrix}\trans \pspace .
\end{align}

\subsection{Fusion Framework}
\label{sec:fusionFrame}
Since the cell resolution is dynamic and, thus, non-uniformly distributed, the APGM requires a specific fusion process.  
Therefore, generic operators are proposed in the following.
Depending on the task, the operator processing the cell content must be specified separately.
The following definition can provide equally sized layers to this fusion operator using a resampling function.
In general, resampling does not necessarily require a memory reallocation; altering the access to the underlying data can be faster in some cases.
Given a required resolution step $r_{req} \in \mathbb{N}$, the fusion yields a layer with the respective resolution if there is a layer with at least the required resolution. 
Otherwise, the highest available resolution is used instead.

With a resampling function $R: \mathcal{L} \times \mathbb{N} \rightarrow \mathcal{L}$, and the type-specific cell fusion operator $\textit{f}_t: \mathcal{L}_t \times \ldots \times \mathcal{L}_t \rightarrow \mathcal{L}_t$, and a set of $n \in \mathbb{N}$ layers $S_l$ the layer fusion function $\textit{F}_{layer}: \mathcal{L}_t \times \ldots \times \mathcal{L}_t \rightarrow \mathcal{L}_t$ is defined by
\begin{align}
    \textit{F}_{\mathrm{layer}}(S_l) = \left\{ \pspace \textit{f}_t \pspace \left(\left\{R\left(l^r_t, r_{\mathrm{fused}}\right) \right\}_{l^r_t \in S_l} \right) \right\}
\end{align}
with $r_{\mathrm{fused}} = \min\left( r_{\mathrm{req}}, \max\left( \left\{r \in \mathbb{N} : \exists l^r_t \in S_l\right\} \right) \right)$.
Examples for $R$ and $\textit{f}_t$ are given in the next section.
Next, given a set of patches $S_p$, the patch fusion function $\textit{F}_{patch}: \mathcal{P} \times \ldots \times \mathcal{P} \rightarrow \mathcal{P}$ is defined by
\begin{align}
    \textit{F}_{\mathrm{patch}}(S_p) = \left\{ \textit{F}_{\mathrm{layer}}(S_{l,t}) : t \in \bigcup_{p \in S_p} \mathcal{I}_p \right\}
\end{align}
with $S_{l,t} = \{l^r_t : \forall p \in S_p , \pspace \exists l^r_t \in p \}$ containing the layer of type $t$ of all patches.
Finally, given a set of grid maps $S_g$, the grid map fusion function $\textit{F}_{grid}: \mathcal{G} \times \ldots \times \mathcal{G} \rightarrow \mathcal{G}$ is defined by
\begin{align}
    \textit{F}_{\mathrm{grid}}(S_g) = \left\{ \textit{F}_{\mathrm{patch}}(S_{(i_x,i_y)}) : (i_x,i_y) \in \bigcup_{g \in S_g} \mathcal{I}_g \right\} \space ,
\end{align}
where $S_{(i_x,i_y)} = \{p_{(i_x,i_y)} : \forall g \in S_g, \pspace \exists p_{(i_x,i_y)} \in g \}$ contains all available patches at index ${(i_x,i_y)}$.

Different cell resolutions and the availability of layers and patches can be considered with the described structure and fusion functions.
With this, the APGM can implement dynamic, external requirements specifying those values.
For example, the authors of~\cite{henning2022situation} propose a method to provide such information depending on the current situation (cf. Section~\ref{sec:relWork}).
As shown in Section~\ref{sec:experiments}, compared to having static grid and cell layouts, the total amount of cells is reduced by realizing such requirements.
Thus, overall memory usage can be improved.
If data are transmitted over links with limited bandwidths~\cite{ushift2022}, the allocated memory size can directly reduce latency and, thus, is of particular importance.
The APGM is exceptionally well-suited for setups as proposed in~\cite{ushift2022} since sensor FOVs are permanently limited to certain directions~\cite{henning2022situation2}.

\subsection{Adaptive Resolution and Layer Resampling}
\label{sec:adapRes}
\begin{figure}[t]
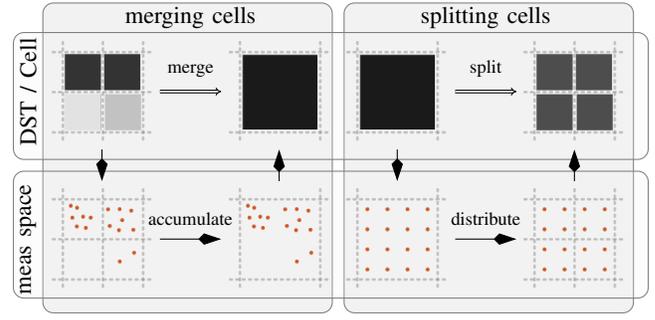

\vspace{0.1cm}
    \centering
    \include{img/split_merge.tex}
    \caption{%
        The process of splitting and merging cells is illustrated.
        For this, the DST information of a cell is transformed into the measurement space using the inverse measurement model.
        In the case of lidar sensors, the measurement space is a point cloud (red dots).
        After fusing the data in the measurement space, the result is transformed into the DST space.
    }
    \label{fig:adapRes}
    \vspace{-0.2cm}
\end{figure}
In this section, we derive a formalism that allows the implementation of the resampling function $R$ required by the previous section.
As mentioned above, only point cloud measurements, e.g., from lidar sensors, are considered here;
And an in-depth definition for camera semantic label with an arbitrary amount of semantic hypotheses is subject to future work.
As described in this section, the idea of merging and splitting cells is exemplarily visualized in Fig.~\ref{fig:adapRes}.
Given $n \in \mathbb{N}$ cells $C = \{c_1, \ldots, c_n\}, c_i \in \mathcal{C}_{occ}$, and a measurement as a set of points $M$, the BBA for $\Omega = \{O, F\}$ representing occupation and free space of each cell can be calculated using an approach proposed by~\cite{richter2022mapping}.
For data consistency, it is crucial that merging these cells yields the same result as calculating the BBA for a merged cell directly.
Analogously, the same applies when cells are split.
As shown in Fig.~\ref{fig:adapRes}, when cells are merged, a free cell does not contradict with occupied cells.
Hence, as shown in the example in~\eqref{eq:dstRuleEx}, the Dempster-Shafer rule of combination is not suited for this task.
Given $M_{c_i}$ as the subset of points $p \in M$ located within a specific cell $c_i \in C$, the evidence of occupancy for this cell $m_{c_i}(O)$ is defined by
\begin{align}
\label{eq:evidence}
    m_{c_i}(O) = 1 - \prod_{p \in M_{c_i}}{\text{Pr}(p \not\to O, c_i)} \pspace ,
\end{align}
where $\text{Pr}(p \not\to O, c_i)$ is the probability, that $p$ is not relevant for the occupancy hypothesis of cell $c_i$~\cite{richter2022mapping}.
Here, \eqref{eq:evidence} is called the grid measurement model.
Next, the merged occupancy evidence $m_{\tilde{c}}(O)$ for the merged grid cell $\tilde{c}$ is calculated as
\begin{subequations}
\label{eq:merge}
\begin{align}
\label{eq:mergeProd}
    m_{\tilde{c}}(O) & = 1 - \prod_{c \in C}{\left(1 - m_{c}(O)\right)} \\
             & = 1 - \prod_{c \in C}{\prod_{p \in M_{c}}{\text{Pr}(p \not\to O, c)}} \\
             & = 1 - \prod_{p \in M_{\tilde{c}}}{\text{Pr}(p \not\to O, c)} \pspace ,
\end{align}
\end{subequations}
where, given the spatial proximity, it is assumed that the sensor measurement model $\text{Pr}(c \pspace | \pspace p)$ is cell invariant.
Generally speaking, merging cells using~\eqref{eq:merge} fuses the information in the measurement space using the inverse of the grid measurement model~\eqref{eq:evidence}.
Without further information about the original point distribution, the process of merging in~\eqref{eq:merge} can only be inverted by distributing the probability $\text{Pr}(p \not\to O, c)$ evenly over the cells.
Hence, the occupancy evidence for the cells $c_i$ after splitting is given by
\begin{align}
\label{eq:split}
\begin{aligned}
    m_{c_i}(O) & = 1 - \left( \prod_{p \in M}{\text{Pr}(p \not\to O, c_i)} \right)^{1 / n} \pspace .
\end{aligned}
\end{align}

Depending on the modeling of free space, it must be considered separately.
With the approach of~\cite{nuss2016random}, the measurement model only considers free space in non-occupied cells.
For this,~\eqref{eq:merge} and \eqref{eq:split} are sufficient to process occupied cells.
In contrast, the authors of~\cite{richter2022mapping} propose determining free space evidence in occupied cells.
Here, a model-specific update considering free space masses is additionally required.
In the case of unoccupied cells, standard fusion operations can be used for merging cells since the free space is considered solely.
Splitting free cells uses a respective inverse fusion operation. 
For example, when median fusion is used for merging cells, splitting can be realized by assigning the value of the original cell to the newly created cells.

\section{EXPERIMENTS}
\label{sec:experiments}
\begin{figure*}
    \centering
    \subfloat[Parking Lot]{%
        \label{fig:apgmPics:parking}%
    	\includegraphics[
    	    width=0.3\textwidth, height=0.3\textwidth
    	]{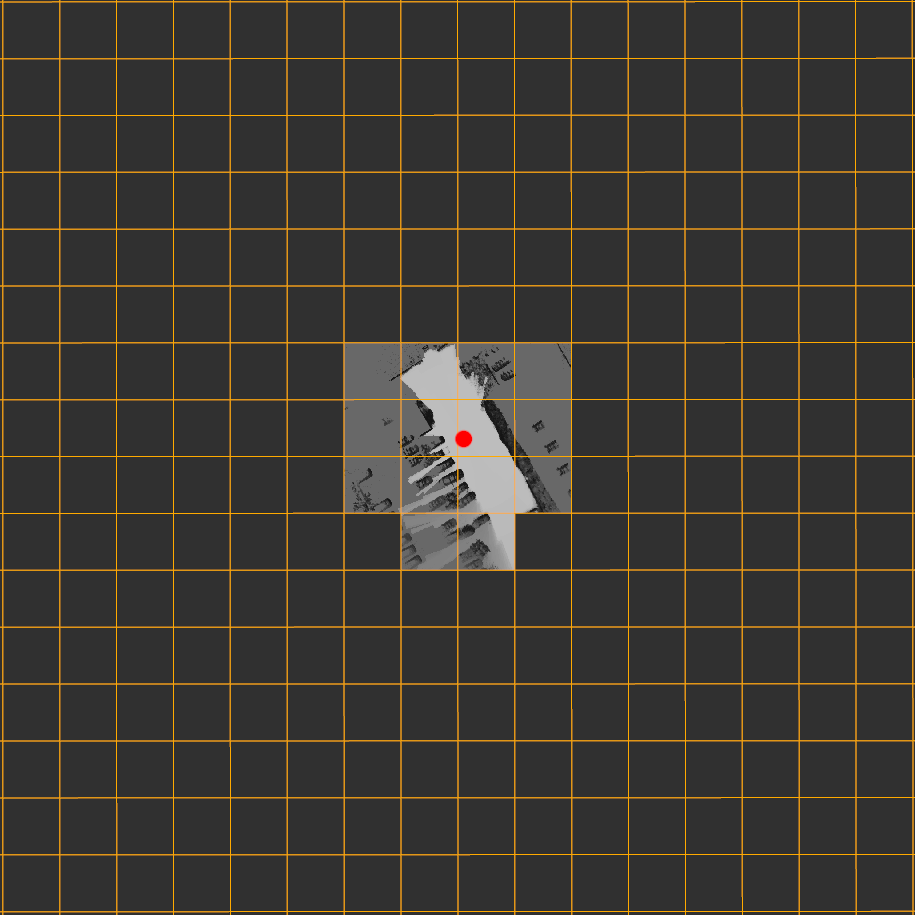}%
    }
    \hfill
    \subfloat[Comparison]{%
        \label{fig:apgmPics:comp}
    	\includegraphics[width=0.3\textwidth, height=0.3\textwidth]{patch_comparison.tikz}%
    }
    \hfill
    \subfloat[Driving]{%
        \label{fig:apgmPics:driving}%
    	\includegraphics[
    	    width=0.3\textwidth, height=0.3\textwidth
    	]{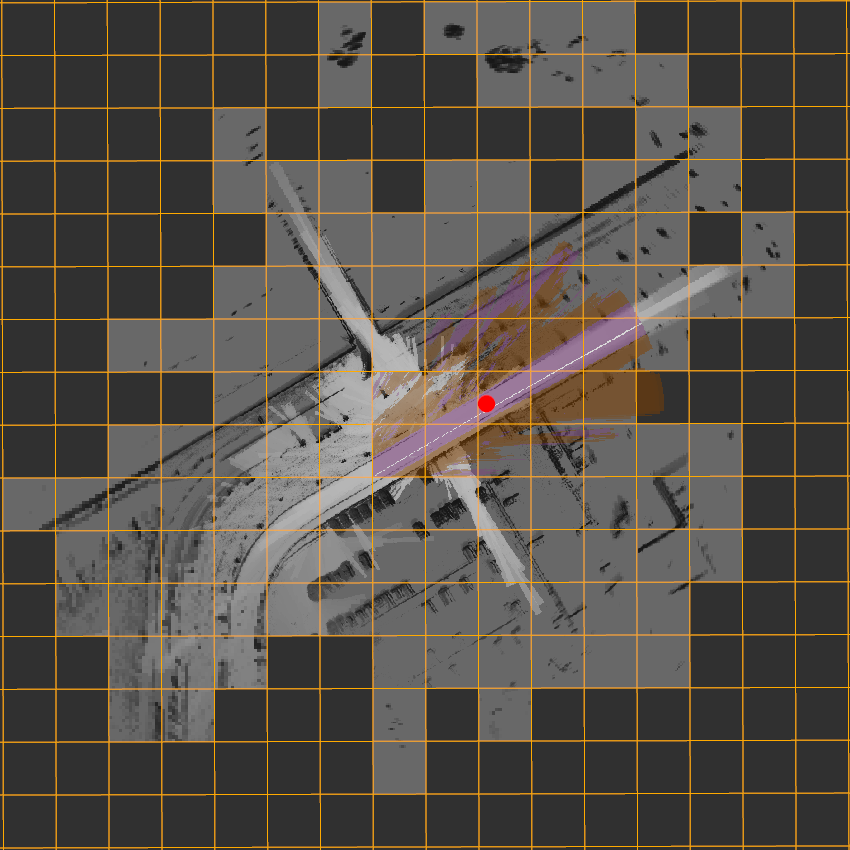}%
    }
    \caption{%
    Three examples of the APGM during the evaluation scenario are shown. 
    Each image shows an area of approx. $\SI{200}{\meter} \times \SI{200}{\meter}$ with the patch division of $\SI{12.8}{\meter}$ in orange and the vehicle position in red.
    \protect\subref{fig:apgmPics:parking} shows the APGM at the parking lot.
    It can be seen that the required horizon of $\SI{20}{\meter}$ is realized.
    \protect\subref{fig:apgmPics:driving} shows the APGM while driving in an urban area.
    Empty patches within the horizon show that patches are successfully removed if no measurements are available.
    For a direct comparison, \protect\subref{fig:apgmPics:comp} shows the occupancy layers of a single patch from both situations, respectively, in more detail.
    The parking lot layer shows more details due to the increased cell resolution.
    }
    \label{fig:apgmPics}
\end{figure*}
This section first describes the scenario used to compare our approach to others.
Afterward, example images of the APGM are given for different points in the scenario.
Finally, the results are presented and discussed, leading to recommendations for using the APGM.

\subsection{Evaluation Scenario}
For evaluation purposes, a reference scenario is defined in this section.
In the scenario, an autonomous vehicle starts from a parking lot.

The first task is to leave the parking lot with a maximum speed of $\SI{15}{\km/\hour}$ and without any topology information available.
Thus, the surrounding environment must be perceived using onboard sensors only.
For this, lidar sensors are used since they are well-suited for unstructured environment perception. 
The cell size of a grid map is selected to be $\SI{10}{\centi\meter}$ at most.
Due to the low speed, a horizon of $\SI{20}{\meter}$ is assumed to be sufficient.
Fig.~\ref{fig:apgmPics:parking} shows an image of the APGM for this part of the scenario.

When leaving the parking lot, the vehicle enters road traffic in an urban environment.
Since road maps are available, grid map information is mainly used for collision avoidance.
Therefore, the cell resolution should approximately reflect the lidar resolution with a lower bound of $\SI{20}{\centi\meter}$.
The horizon must be at least $\SI{100}{\meter}$, corresponding to approx. $\SI{7}{\second}$ of driving with $\SI{50}{\km/\hour}$.
Additionally, a front-facing camera must be used since curbs and road markings are hard to detect with lidars only.
However, ground projection errors of camera information quickly increase with increasing distance.
Thus, the required horizon for camera semantic information is only $\SI{40}{\meter}$, which still allows the vehicle to react correctly on detections of curbs and road markings.
Two exemplary images of the vehicle leaving the parking lot and driving on the road are shown in Fig.~\ref{fig:apgmPics:parking} and~\ref{fig:apgmPics:driving}, respectively.

Last, the vehicle enters another parking lot, and the scenario ends with the vehicle parked.
For this part, the same grid map requirements as for the parking before are used.

\subsection{Evaluation Setup}
The vehicle used for evaluation is an autonomous vehicle with multiple lidars and cameras.
As required, the $\SI{360}{\degree}$ FOV are covered by two lidars (Hesai Pandar 64).
Additionally, a front-facing $\SI{3}{MP}$ camera with a semantic segmentation module is available.
Lidar information is encoded as occupancy measurement represented by a two dimensional BBA per cell with $T = \{O,F\}$.
Camera information is encoded as semantic measurement represented by a four-dimensional BBA per cell with $T = \{Road, Marking, Blocked, Unknown\}$.

Our approach is compared to the approaches of~\cite{buerkle2020efficient} and~\cite{wellhausen2021efficient}.
The comparison comprises the number of cells and the required memory size of each approach.
For this comparison, the "non-uniform [$\SI{20}{\meter}$:$\SI{60}{\meter}$:$\SI{100}{\meter}$]" setting from~\cite{buerkle2020efficient} is the best of the presented settings which fits the requirements and is, thus, used as a reference.
The APGM is configured to match the same cell resolution and horizon specifications.
Since the~\cite{buerkle2020efficient} also estimates velocities, a fair comparison of run times is impossible.
However, average fusion execution times are determined for the APGM as a reference.
The evaluation was run on a computer containing an ADM\,Ryzen\textsuperscript{TM}\,Threadripper\textsuperscript{TM}\,3970X CPU and 64GB of DDR4 RAM.

\subsection{Results}
\begin{figure*}[ht]
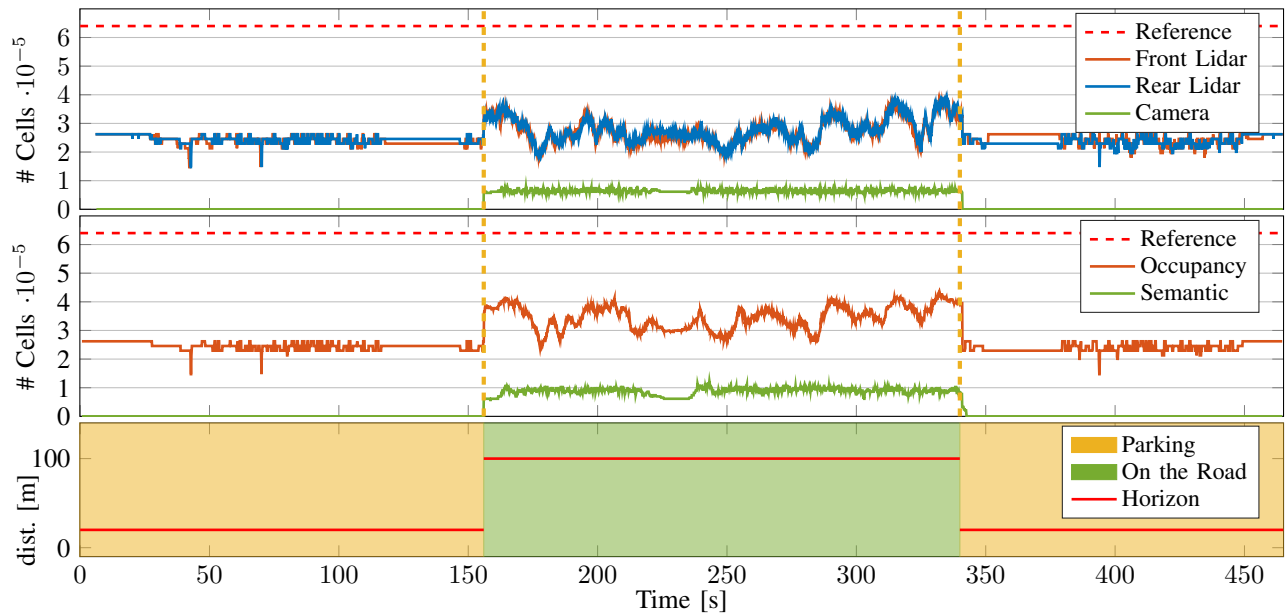

\vspace{0.3cm}
    \centering
    \begin{tikzpicture}[baseline=(current bounding box.center)]

    \begin{groupplot}[
        group style={
            group size=1 by 3,
            x descriptions at=edge bottom,
            vertical sep=2.5pt,
            xlabels at=edge bottom
        },
        scale only axis,
        legend cell align={left},
        xlabel style={yshift=1.5mm},
        xlabel={Time [s]}
    ]

    \nextgroupplot[
        width=0.9\linewidth,
        height = 0.15\linewidth,
        scale only axis,
        xmin=-0.1,
        xmax=465.1,
        ymin=0,
        ymax=700000,
        ytick={0,100000,200000,300000,400000,500000,600000},
        ytick scale label code/.code={},
        ylabel style={yshift=0.6mm},
        ylabel={\# Cells $\cdot 10^{-5}$},
        ymajorgrids,
        axis background/.style={fill=white},
        legend style={font = \large, legend cell align=left, align=left, draw=white!15!black,nodes={scale=0.75, transform shape}}
    ]
    \addplot [dashed, color=red, line width=1pt] table [col sep=comma] 
    {
        0, 640000
        465, 640000
    };
    \addlegendentry{Reference}
    \input{img/raw_datapoints/ul1500-lidar_front.txt}
    \addlegendentry{Front Lidar}
    \input{img/raw_datapoints/ul1500-lidar_rear.txt}
    \addlegendentry{Rear Lidar}
    \input{img/raw_datapoints/ul1500-camera_front_left_stereo.txt}
    \addlegendentry{Camera}

    \addplot +[mark=none, color=matlabYellow, dashed, line width=1.6pt] coordinates {(156, -1) (156, 750000)};
    \addplot +[mark=none, color=matlabYellow, dashed, line width=1.6pt] coordinates {(340, -1) (340, 750000)};
    
    \nextgroupplot[
        width=0.9\linewidth,
        height = 0.15\linewidth,
        scale only axis,
        xmin=-0.1,
        xmax=465.1,
        ymin=0,
        ymax=700000,
        ytick={0,100000,200000,300000,400000,500000,600000},
        yticklabels={{0},{1}, {2},{3},{4},{5},{6}},
        ytick scale label code/.code={},
        ylabel style={yshift=0.6mm},
        ylabel={\# Cells $\cdot 10^{-5}$},
        ymajorgrids,
        axis background/.style={fill=white},
        legend style={font = \large, legend cell align=left, align=left, draw=white!15!black,nodes={scale=0.75, transform shape}}
    ]    
    \addplot [dashed, color=red, line width=1pt] table [col sep=comma] 
    {
        0, 640000
        465, 640000
    };
    \addlegendentry{Reference}
    \input{img/raw_datapoints/ul1500-odom-occ.txt}
    \addlegendentry{Occupancy}
    \input{img/raw_datapoints/ul1500-odom-sem.txt}
    \addlegendentry{Semantic}
    
    \addplot +[mark=none, color=matlabYellow, dashed, line width=1.6pt] coordinates {(156, -1) (156, 700000)};
    \addplot +[mark=none, color=matlabYellow, dashed, line width=1.6pt] coordinates {(340, -1) (340, 700000)};

    \nextgroupplot[%
        width=0.9\linewidth,
        height = 0.1\linewidth,
        scale only axis,
        scaled y ticks = false,
        xmin=-0.1,
        xmax=465.1,
        ymin=-10,
        ymax=140,
        ytick={0, 100},
        yticklabels={{0},{100}},
        ylabel style={yshift=-3.5mm, xshift=-4mm},
        ylabel={dist. [m]},
        axis background/.style={fill=white},
        legend style={font = \large, legend cell align=left, align=left, draw=white!15!black,nodes={scale=0.75, transform shape}}
    ]
    
    \addlegendimage{area legend, color=matlabYellow, fill=matlabYellow}
    \addlegendentry{Parking}    
    \addlegendimage{area legend, color=matlabGreen, fill=matlabGreen}
    \addlegendentry{On the Road}

    \addplot [draw=matlabYellow,fill=matlabYellow, semitransparent,forget plot]
    (-1,-11) rectangle (156, 140);
    \addplot [draw=matlabYellow,fill=matlabYellow, semitransparent,forget plot]
    (340,-11) rectangle (466, 140);
    \addplot [draw=matlabGreen,fill=matlabGreen, semitransparent,forget plot]
    (156,-11) rectangle (340, 140);

    \addplot [color=red, line width=1pt,forget plot] 
        table [col sep=comma]  
    {
        0, 20 
        156, 20 
    };
    \addplot [color=red, line width=1pt,forget plot] 
        table [col sep=comma]  
    {
        156, 100 
        340, 100 
    };
    \addplot [color=red, line width=1pt] 
        table [col sep=comma]  
    {
        340, 20 
        465, 20
    };
    \addlegendentry{Horizon}

    \end{groupplot}
    
\end{tikzpicture}%
    \caption{
        Here, the amount of cells of the APGM during the evaluation scenario is shown.
        The upper plot illustrates the number of cells for occupancy measurements of both lidars (red and blue) and the semantic measurement of the camera (green).
        Additionally, the constant amount of cells of the reference approach~\cite{buerkle2020efficient} is shown (dashed red).
        Dashed yellow lines mark the times of requirement changes.
        In the middle, the number of occupancy (red) and semantic (green) cells are plotted after the fusion step.
        The lower graph shows the required horizon distance, and the background further describes the current situation.
        For illustration reasons, the number of cells for the approach of~\cite{wellhausen2021efficient} is not shown. In comparison, the amount of cells equals ours while parking; however, it is at least four times higher while driving on the road.
    }
    \label{fig:cell_count}
\end{figure*}
The number of cells for each layer and the required horizon of the evaluation run is plotted in Fig~\ref{fig:cell_count}.
For comparison, the constant amount of cells required for the approach of~\cite{buerkle2020efficient} is shown as a reference.
It shall be mentioned that the amount of cells is required per measurement grid; thus, values for the APGMs are separately compared to the reference. 
However, due to the lack of resolution changes, the number of cells for the approach of~\cite{wellhausen2021efficient} are equal to ours while parking and at least four times higher than ours while on roads.
For lucidity reasons, they are not added to the plot.
The average fusion execution times for the APGM using eight threads are $\SI{1.8}{\milli\second}$ and $\SI{2.6}{\milli\second}$ for the parking and on roads, respectively.

As shown in Fig.~\ref{fig:apgmPics} and~\ref{fig:cell_count}, our approach can realize external requirements, including horizon distances, cell resolutions, and data availability.
Due to the lower required cell resolution, the amount of cells is not increasing proportionally with the horizon distance, e.g., when entering the road.
The number of occupancy cells in the fused grid map is generally higher than the amounts for a single lidar measurement grid.
This is a result of the partly overlapping FOVs of both sensors.
Over the whole scenario, the amount of occupancy grid cells is significantly lower than the reference number of~\cite{buerkle2020efficient}.
On average, the fused grid has $243k$ and $345k$ occupancy cells in the parking lot and on the road, respectively, and $88k$ semantic cells on the road.
Thus, the memory efficiency is improved by a factor of $7.9$ in the parking lot and $3.7$ on the road.
The amount of patches of~\cite{wellhausen2021efficient} is equal to the APGM.
However, when driving in an urban situation, their cell resolution and the number of cells is higher, and each cell allocates more memory.
Hence, the memory efficiency is improved by a factor $\geq 3.0$ while parking and $\geq 8.0$ while driving on the road.

\begin{figure}[t]
\vspace{0.2cm}
    \centering
    \includegraphics[width=\linewidth]{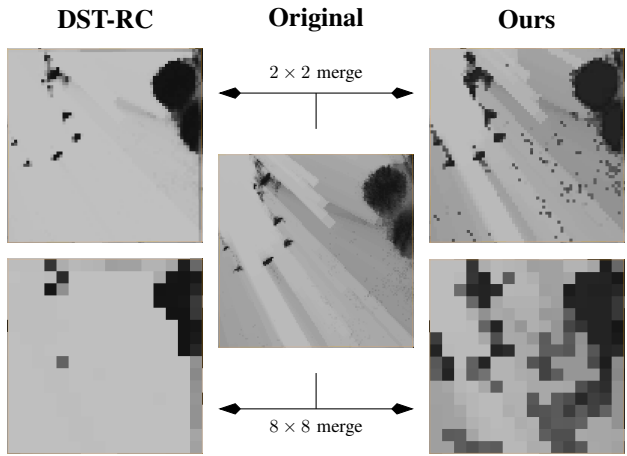}
    \caption{%
        A visual example of our proposed resampling approach compared to the DST-RC is shown.
        In the middle, an occupancy layer with a cell size of $\SI{10}{\centi\meter} \times \SI{10}{\centi\meter}$ prior to resamping is displayed as reference.
        On the left and right, the results of merging $2 \times 2$ and $8 \times 8$ cells with the DST-RC and our resampling method are depicted, respectively.
    }
    \label{fig:changes}
\end{figure}
Further, an example of our proposed approach for resolution resampling compared to DST-RC resampling is shown in Fig.~\ref{fig:changes}.
Occupied areas tend to shrink with the DST-RC due to conflicts with the surrounding free space.
For the $8 \times 8$ merging, the DST-RC resampling loses track of occupied areas.
This effect could lead to collisions when driving in tight spaces.
In contrast, our approach successfully keeps cells occupied when any of the merged cells was occupied before.
Additionally, the $2 \times 2$ merge using our approach preserves the uncertainty information of cells, where the DST-RC method delivers overly certain free space estimations.
This effect can be seen, in the lower right of the $2 \times 2$ merge images in Fig.~\ref{fig:changes}.

Summarizing the results, the APGM can transfer requirement facilitation into improved memory consumption and is, thus, more efficient than other approaches.
Additionally, when using one of the compared approaches, a trade-off decision between different situations has to be made, whereas our approach performs best in all situations.
Given the short fusion execution times, our approach allows high update rates of over $\SI{100}{\hertz}$; thus, multiple sensors can be used together in real-time.
The results also show that the dedicated consideration of cell resampling is essential, and standard fusion operations, like the DST-RC, cannot be used for adaptive resolution processing.
\section{CONCLUSION}
\label{sec:conclusion}

In this work, we have proposed the Adaptive Patched Grid Map (APGM), which models the unstructured environment depending on dynamically changing situational requirements.
This is enabled by our new fusion framework, which allows using Dempster-Shafer theory of evidence (DST) fusion operators on cell level.
It considers various cell resolutions by resampling, i.e., merging and splitting cells.
Our proposed spatial cell fusion fills the gap of available DST fusion operators for the resulting resampling requirements.

We showed the effectiveness of our approach using real-world data recorded from an autonomous vehicle.
The APGM significantly improved memory efficiency compared to other approaches while still fulfilling all situational requirements.
At the same time, low execution times enable high update rates.
Thus, our approach contributes to situation-aware adaptive perception methods, allowing for resource savings.

For future work, we plan to add an adaptive velocity estimation and the consideration of sensor trustworthiness.

{\small
\bibliographystyle{IEEEtran}
\bibliography{bib.bib}
}

\end{document}